\documentclass[10pt,twocolumn,letterpaper]{article}

\usepackage[pagenumbers]{cvpr} 

\definecolor{cvprblue}{rgb}{0.21,0.49,0.74}
\usepackage[pagebackref,breaklinks,colorlinks,allcolors=cvprblue]{hyperref}

\title{A Dataset for Dynamic Human Preferences for Vision Language Models}

\author{Hannah Gao\\
MIT\\
Cambridge, MA\\
{\tt\small hanngao@mit.edu}
\and
Dylan Hadfield-Menell\\
MIT\\
Cambridge, MA\\
{\tt\small dylanhm@mit.edu}
\and
Rachel Ma\\
MIT\\
Cambridge, MA\\
{\tt\small rachelm8@mit.edu}
}

\begin{document}
\maketitle
\begin{abstract}
Given the increased adoption of Vision Language Models (VLMs) in human-interactive settings, it is important that we evaluate how well these models can adapt to real-time preferences for different users. While an increasing number of vision-language benchmarks have recently been introduced, they focus largely on evaluating static capabilities and generally-held preferences learned from extensive training data. This work introduces a new benchmark for evaluating the ability of VLMs to understand dynamic human-preferences, i.e. preferences that are passed in-context at inference time. We provide an automated pipeline for generating this benchmark with variations on image dependence, a dynamic multi-modal human-preference dataset, and evaluations of state-of-the-art models on the novel benchmark.
\end{abstract}    
\section{Introduction}
\label{sec:intro}
In recent years, Vision Language Models (VLMs) have become increasingly popular, not only demonstrating impressive generative capabilities on multi-modal inputs, but also serving as powerful perception and knowledge backbones for vision language action models (VLAs) ~\cite{kim2024openvla, intelligence2025vision, geminiroboticsteam2025geminiroboticsbringingai}.  From chatbot assistants ~\cite{openai2024gpt4technicalreport} to VLM-powered robotics, the wide-ranging applications of VLMs in interactive settings makes it critical that we have proper datasets and evaluation benchmarks so that we can 1) examine the ability of these models to adhere to preferences raised by humans and 2) fine-tune VLMs towards this goal where needed. 

A number of multimodal benchmark datasets for evaluating VLMs on a wide range of vision-language tasks have been introduced ~\cite{liu2024mmbench, yue2024mmmu, antol2015vqa, goyal2017making, li2023seed, lin2014microsoft}. However, these datasets tend to focus on knowledge-extraction-based tasks
such as scene understanding and object recognition, rather than tasks that involve personalization to the user. 

It is important that we also evaluate these models on human preferences in a more context-dependent, personalized sense. For example, to make VLMs more helpful for individual users, they should be able to adapt to real-time preferences passed in by the user that may or may not align with a ``typical" or popularly-held preference. We will call these ``dynamic" human preferences. For example, in arranging books on a bookshelf, some individuals may prefer the books to be arranged by color while others may prefer to arrange by alphabetical order. Sometimes, real-time preferences could even be serious if not acknowledged, such as preferences related to allergies or other medical conditions.

In this paper, we attempt to fill this gap by providing the following contributions. First, we introduce a preliminary evaluation benchmark for visual reasoning about human preferences. Rather than evaluating models on their adherence to generally-held preferences (e.g. overall helpfulness), however, we evaluate human preferences in the ``dynamic" sense, where the preference is user-specific and is passed in at inference time. We also propose a generative pipeline for producing the dataset, including a method for automatically generating visually-dependent prompts by introducing image bounding boxes. Finally, we evaluate the dataset on various closed-source and open-source VLMs.


\section{Related Works}

Recently, VLMs have seen great advances, achieving impressive results on both vision and language tasks ~\cite{guan2024loc, zhao2024vlm, nacson2025docvlm}.
However, despite their growing competence, VLMs have been found to struggle in cross-modal applications, particularly ones with heavy reliance on the visual domain, including visual perception of geometric shapes ~\cite{kamoi2024visonlyqa} and visual arithmetic ~\cite{huang2025vision}, necessitating proper benchmarking of these models. As such, an increasing number of benchmarks have been created to assess VLM cross-modal capabilities, with increasing breadth and difficulty. 

Prior benchmarks, including several captioning datasets (e.g. Microsoft COCO Caption~\cite{chen2015microsoft} and NoCaps~\cite{agrawal2019nocaps}) and VQA datasets (e.g. GQA~\cite{hudson2019gqa} and Q-BENCH+~\cite{zhang2024q}) are known to be rather narrow in scope or rigid in their evaluations ~\cite{liu2024mmbench}. In recent years, more general and comprehensive benchmarks for VLMs have emerged, including MMBench and MMMU ~\cite{liu2024mmbench, yue2024mmmu, li2024omnibench, kamoi2024visonlyqa, guan2024hallusionbench}. However, those benchmarks focus on evaluating models on tasks based on static knowledge gained from extensive training. They do not capture well the personalization capabilities of generative models, i.e. the ability of models to adapt to real-time preferences of a human user which may or may not coincide with static preferences that dominate in the training data.

A few benchmarks have looked at human preferences, but not in a dynamic, in-context sense, often defining preference as a generalizable tendency to select one model output over another ~\cite{wu2023human} or as the overall helpfulness of the model's responses ~\cite{ji2023beavertails, ethayarajh2021understanding, zhang2024learning}. For example, WildVision ~\cite{lu2024wildvision}, a large-scale vision-language dataset, discusses preferences in terms of preferred answers between two model responses. These datasets capture widely-held static preferences over the desired traits of model responses rather than evaluate the model's ability to personalize responses to dynamic preferences passed in at inference time. 
\section{Methods}

\subsection{Visual Human Preference Dataset Description}

\begin{figure}[t]
  \centering
    \includegraphics[width=1.0\linewidth]{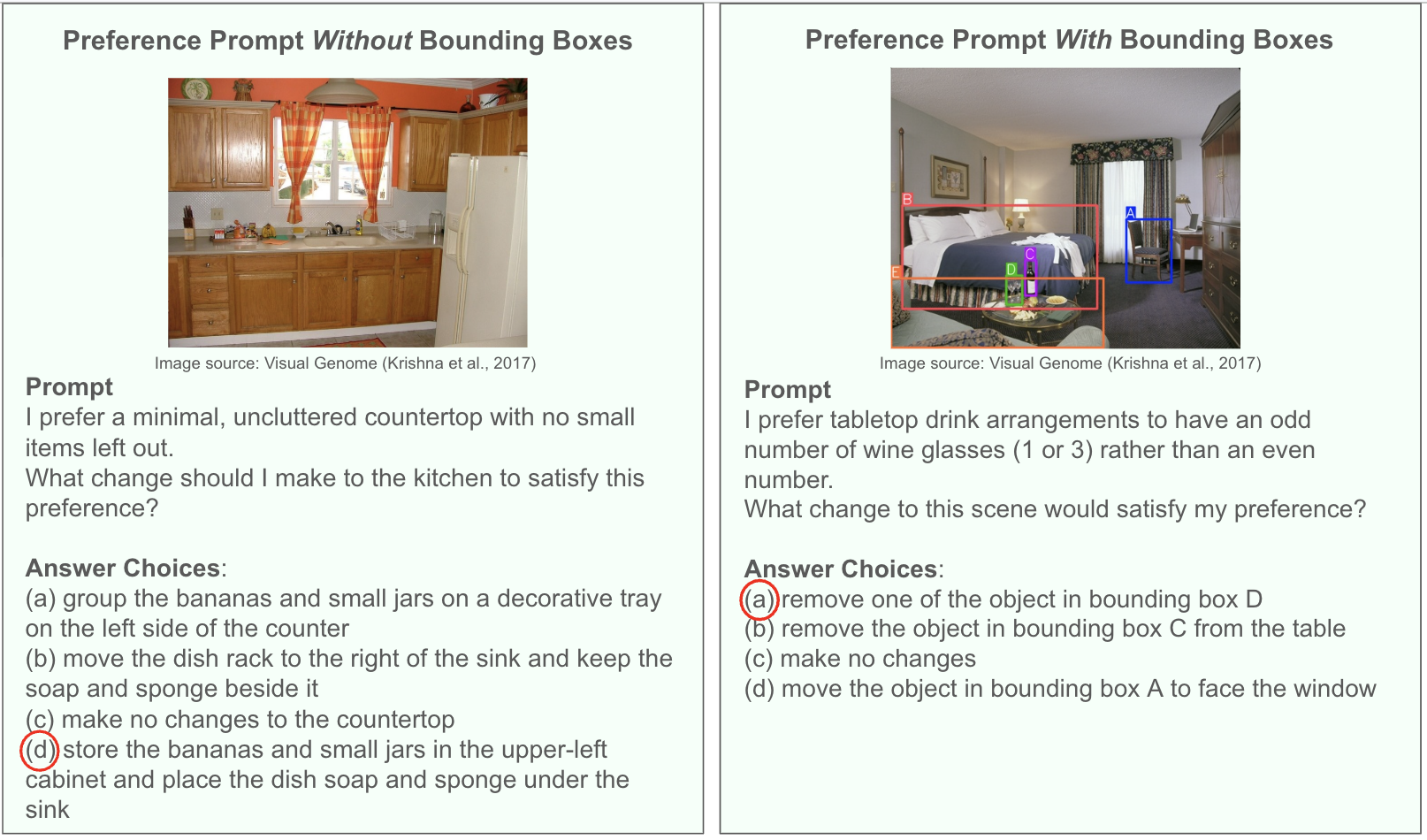}
   \caption{Example of a prompt without bounding boxes (left) and with bounding boxes (right), with human-determined ground-truths circled. Prompt images are from \href{https://homes.cs.washington.edu/~ranjay/visualgenome/index.html}{Visual Genome} (introduced by Krishna et al., 2017 \cite{krishna2017visual}; licensed under \href{https://creativecommons.org/licenses/by/4.0/}{CC BY 4.0}), modified in this work to include bounding boxes.}
   \label{fig:example_prompt}
\end{figure}

To provide both visual dependencies and dynamic human preferences, we present a dataset that consists of an image, a stated human preference, a relevant question, and four answer choices. All answer choices are actionable (i.e. begin with a verb) making our dataset particularly suitable for agentic tasks. See \Cref{fig:example_prompt} for example prompts.
We design an automated pipeline to generate this data in 4 variations based on combinations of the prompt's image-dependency and how typical the preference is: 0 = (low dependence, typical), 1 = (high dependence, typical), 2 = (low dependence, atypical), 3 = (high dependence, atypical). Here, image-dependency refers to the degree to which answering the text prompt requires information from the image.

We also include a variation of this dataset that references image bounding boxes to more strongly enforce high image dependency of the dataset.  This data includes the same components as the non-bounding-box data except that the images for the prompts include bounding boxes around objects, and answer choices reference objects by their bounding box letter in the image (see \cref{fig:example_prompt}).

\subsection{Automated Data Generation Pipeline}

\begin{figure}[t]
  \centering
    \includegraphics[width=0.8\linewidth]{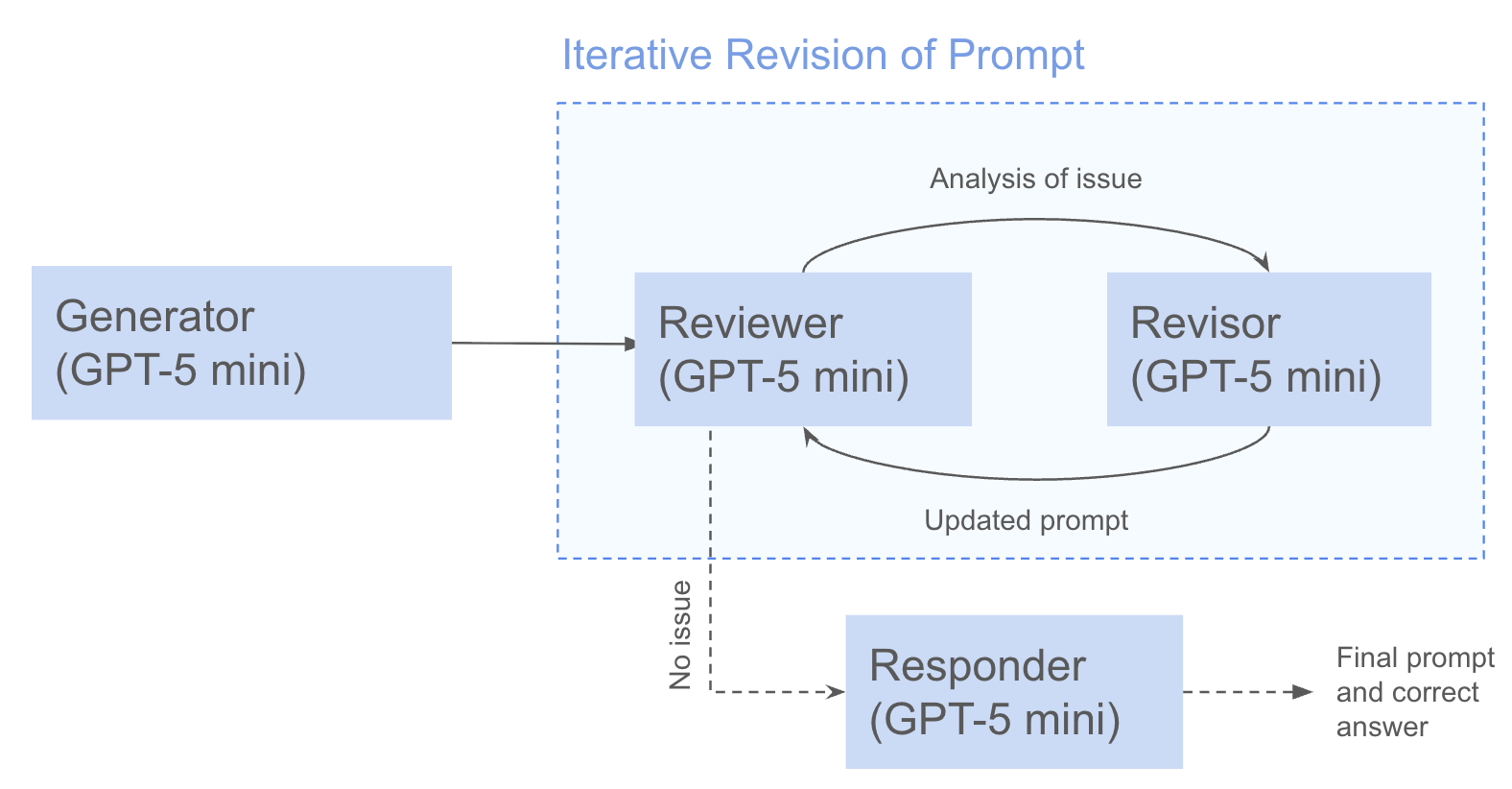}
   \caption{Overview of the VLM-powered data generation and iterative revision pipeline.}
   \label{fig:data_gen_pipeline}
\end{figure}

We automate the generation and revision of our data using a pipeline consisting of these 4 components: generator, reviewer, revisor, responder. Prior works have found success using iterative self-revision pipelines for LLM outputs \cite{hu2024evoke, madaan2023self}. \cref{fig:data_gen_pipeline} provides an overview of the pipeline setup.
\\\\
\textbf{1. Generator}: The generator creates the initial prompt, given an image. It outputs a scene description, a human preference, a related question, and answer choices a-d (where all answer choices should be actionable). Depending on the category, we include additional specifications such as including prompting for atypical preferences for categories 2 and 3, or requesting higher image dependency for categories 1 and 3.  
\\\\
\textbf{2. Reviewer}: The reviewer generates a review of the generated prompt. It is given a list of possible issues: wrong scene description, poor preference construction, no correct answer choices, multiple correct answer choices, and non-actionable answer choices. Low image dependency is additionally included as an issue for categories 1 and 3. The reviewer analyzes possible issues with the prompt and can pick an issue from the list or respond that there is no issue, at which point the responder is called.
\\\\
\textbf{3. Revisor}: Based on the issue picked by the reviewer, we provide the revisor with the prompt and a suggested possible fix. The revisor must then output its proposed fix to the identified issue and a revised version of the prompt.
\\\\
\textbf{4. Responder}: The responder must answer the prompt with a single letter correct answer choice based on the prompt.

\subsection{Bounding Boxes Variation of the Pipeline}
The bounding box variation of the dataset is generated using a slight modification to the pipeline. We use Yolo ~\cite{redmon2016you}, an object-detection model, to both generate up to 5 bounding boxes around objects in the images and provide classifications of those objects. We filter out bounding boxes surrounding people and label each bounding box in the image as a letter from A to E.

We then leverage the pipeline above to generate prompts, providing the original image (without bounding boxes) and a list of the bounding-box object classes in each image, instructing the model to mention the objects in the list in its answer choices. By providing the model with a toolkit of objects it can talk about, as well as the original image, we prevent poor prompt quality as a direct result of the VLM’s misunderstanding of bounding boxes. We then post-process the prompts produced by the pipeline by replacing each of the tagged objects with their classification.

We use a VLM to minimally revise mappings from boxes to categories in case some object class could ambiguously refer to multiple objects in the image. We additionally discard images with multiple bounding boxes of the same class and ask the model to remove bounding boxes that are completely misclassified or that span a large region of the image or contain many smaller bounding boxes within. Finally, we discard any prompt whose image contain less than 2 bounding boxes after the processing described.

\section{Experiments}
\subsection{Models and Datasets}
We use GPT-5-mini~\cite{gpt-5-mini} (at the default reasoning level) as the VLM for all stages of the data-generation pipeline. All images in the dataset are sourced from Visual Genome~\cite{krishna2017visual}, modified to include bounding boxes.

\subsection{Answer Matching}
Matching VLM outputs with ground-truth answer choices is non-trivial as VLM may output extraneous text or format answers in unexpected ways. Inspired by~\cite{liu2024mmbench}, which found success in using a LLM judge to extract the multiple-choice answer from a VLM output, we take a similar approach. We first try a direct matching of the VLM output to the correct letter answer choice. If that fails, we use OpenAI's o4-mini~\cite{gpt-o4-mini} as the answer extractor and ask it to try to match the VLMs output to an answer choice as a fallback.

\subsection{Ground Truth via Human Surveys}
To collect ground-truth human-preference answers for our dataset, we recruit participants on Prolific ~\cite{Prolific} to respond to Google Forms containing a subset of our prompts. In particular, we batch 84 non-bounding-box prompts (4 categories for each of 21 images) into 7 forms and 44 bounding-box prompts (2 categories for each of 22 images) into 4 forms. Each prompt consists of an image, preference, question, and answer choices. For each prompt, we ask 5 participants to select the single best answer choice to the prompt, with an additional ``other" option to indicate an ill-formed prompt or lack of a single correct answer choice. Participants are also asked to rate the question difficulty on a scale from 1 (very easy) to 5 (very difficult) and provide a brief justification for their answer choice. 

To determine ground-truth answers for prompts, we mark any answer choice (excluding the ``other" option) receiving a majority selection among form respondents as the correct answer. Any prompts not receiving convergence to a majority answer are discarded, leaving 76 and 36 prompts in our non-bounding-box and bounding-box datasets respectively that have human-determined ground truths.

Overall, we find that the answers to these prompts produced by the pipeline's responder model (GPT-5-mini) align fairly well with the ground-truth values answer choices. Among prompts with human-labeled ground-truths, GPT-5-mini and survey-majority agreed on answers to 94.7\% of prompts in the dataset without bounding boxes and 88.9\% of prompts in the dataset with bounding boxes.

\subsection{Evaluation Results}

We evaluated several models on our human-labeled dataset, both the non-bounding-box and bounding-box variants. We use ground-truth values collected from the Prolific surveys (described in the previous section) as the correct answers. Closed-source models include GPT-4.1 \cite{openai2024gpt4technicalreport, gpt-4.1} and GPT-5-mini \cite{gpt-5-mini} (with ``medium"-level reasoning); open source models include InstructBLIP-7B~\cite{instructblip, dai2023instructblip}, LLaVA-NeXT-7B~\cite{liu2023improved, liu2024llavanext, liu2023llava}, Qwen3-VL-8B, and Qwen3-VL-32B~\cite{qwen3technicalreport, Qwen-VL}. The results are shown in \cref{tab:model_accuracy_bb} and \cref{tab:model_accuracy_reg}

\begin{table}[h]
  \caption{Accuracy of different models on typical and atypical preference tasks for the bounding-box variant of the dataset.}
  \label{tab:model_accuracy_bb}
  \centering
  \begin{tabular}{@{}lccc@{}}
    \toprule
    Model & \shortstack{Overall \\ Accuracy} & \shortstack{Typical \\Preference \\ Accuracy} & \shortstack{Atypical \\Preference \\ Accuracy} \\
    \midrule
    GPT-4.1 & 86.1\% & 68.8\% & 100.0\% \\
    GPT-5 mini & 88.9\% & 75.0\% & 100.0\% \\
    LLaVA-NeXT-7B & 63.9\% & 68.8\% & 60.0\% \\
    Qwen3-VL-8B-\\Instruct & 75.0\% & 68.8\% & 80.0\% \\
    Qwen3-VL-32B-\\Instruct & 80.6\% & 75.0\% & 85.0\% \\
    InstructBLIP-7B & 47.2\% & 56.3\% & 40.0\% \\
    \bottomrule
\end{tabular}
\end{table}

\begin{table}[h]
  \caption{Accuracy of different models on typical and atypical preference tasks for the non-bounding-box variant of the dataset.}
  \label{tab:model_accuracy_reg}
  \centering
  \begin{tabular}{@{}lccc@{}}
    \toprule
    Model & \shortstack{Overall \\ Accuracy} & \shortstack{Typical \\Preference \\ Accuracy} & \shortstack{Atypical \\Preference \\ Accuracy} \\
    \midrule
    GPT-4.1 & 92.1\% & 91.9\% & 92.3\% \\
    GPT-5 mini & 94.7\% & 91.9\% & 97.4\% \\
    LLaVA-NeXT-7B & 80.3\% & 75.7\% & 84.6\% \\
    Qwen3-VL-8B-\\Instruct & 89.5\% & 86.5\% & 92.3\% \\
    Qwen3-VL-32B-\\Instruct & 90.8\% & 91.9\% & 89.7\% \\
    InstructBLIP-7B & 43.4\% & 40.5\% & 46.2\% \\
    \bottomrule
    \end{tabular}
\end{table}

In terms of overall accuracy, we find that the closed-source models outperform the open-source models on both dataset variants, with the performance gap generally more pronounced for the bounding-box variant among the high-performing models. Except for InstructBLIP, models generally perform worse (in terms of overall accuracy) on the bounding-box variant than on the non-bounding-box variant, with none of the models exceeding 75\% accuracy on the typical preferences of the bounding-box variant. A possible explanation for the ubiquitous poor performance on bounding-box prompts could be a lack of representation of bounding-box images in the training data of these models, suggesting the need for more bounding-box representation and overall greater image-reliance in vision-language datasets. We also observe that Qwen3-VL-32B-Instruct consistently outperforms Qwen3-VL-8B-Instruct in all accuracies, except on atypical preferences on non-bounding-box prompts, suggesting that model size may be related to visual human-preference reasoning abilities. 

These results have important implications for downstream applications of open-source VLMs such as VLAs, where a poor ability to understand dynamic human preferences in the backbone VLM may have ramifications for the VLA's ability to act according to real-time user preferences. 

Interestingly, models tended to perform better on the atypical preference categories (i.e. categories 2 and 3) than on the typical preference categories. Human survey responders also tended to rate atypical preference category prompts as less difficult (see \cref{tab:difficult_ratings}). A possible explanation is that atypical categories tended to have more contrived and detailed preferences and option choices, making it easier for the responder to identify the correct answer without deep reasoning or heavy image reliance. We also observe that for the dataset without bounding boxes, participants found the high-dependency category more challenging than the low-dependency category, on average (see \cref{tab:difficult_ratings}). 

\begin{table}
  \caption{Average human-evaluator prompt difficulty-ratings (out of 5) for typical vs. atypical and low vs. high dependency categories. Bounding-box prompts are considered high-dependency.}
  \label{tab:difficult_ratings}
  \centering
  \begin{tabular}{@{}lcccc@{}}
    \toprule
    Dataset Variant & \shortstack{Typical \\ Pref.} & \shortstack{Atypical \\ Pref.} & \shortstack{Low \\ Dep.} & \shortstack{High \\ Dep.} \\
    \midrule
    Bounding-box & 1.48 & 1.44 & -- & {1.46} \\
    Non-bounding-box & 1.54 & 1.42 & 1.36 & 1.59 \\
    \bottomrule
  \end{tabular}
\end{table}

\section{Conclusion}

In this work, we propose a benchmark dataset for evaluating VLMs on dynamic human preferences in a cross-modal setting. We provide a more context-dependent and personalized treatment of "preference" evaluation than do previous works, and we introduce an automated pipeline for generating the benchmark. 

This work helps elucidate the current state of VLM dynamic human preference understanding in a vision-language context, which is important given the many downstream human-interactive applications of VLMs. Evaluating state-of-the-art open-source and closed-source VLMs on the benchmark, we find that, generally, open-source models lag closed-source models on dynamic human preference understanding, suggesting that downstream applications of open-source VLMs (e.g. VLAs) may also benefit from similar benchmarking. We observe that nearly all models tended to struggle more on prompts consisting of bounding-boxes in the images, suggesting a necessity for greater representation of bounding boxes in vision-language datasets to equip VLMs for understanding these contexts. 

We acknowledge that this dataset is still an early-stage benchmark and is limited in size, and we hope that future work can explore an expansion of this benchmark. 

\section{Data Attribution}

Images in this benchmark dataset are from the \href{https://homes.cs.washington.edu/~ranjay/visualgenome/index.html}{Visual Genome Dataset} (Version 1.2), licensed under \href{https://creativecommons.org/licenses/by/4.0/}{Creative Commons Attribution 4.0 International License} and introduced by Krishna et al. (2017) ~\cite{krishna2017visual}. We made minor modifications to some images by adding bounding boxes. 

\section{Ethics Statement}

Ground-truth answers for our dataset were collected via Prolific Surveys in which participants responded to a fixed human-preference prompt given an image. Participants were compensated a fair amount per survey recommended by Prolific. Surveys did not collect personal information and involved minimal risk to participants. This study is considered IRB Exempt.

\section{Acknowledgements}
This project was funded through a gift from Effective Giving.
{
    \small
    \bibliographystyle{ieeenat_fullname}
    \bibliography{main}
}

\end{document}